\documentclass[10pt,twocolumn,letterpaper]{article}

\PassOptionsToPackage{table}{xcolor}

\usepackage{cvpr}              

\definecolor{cvprblue}{rgb}{0.21,0.49,0.74}
\usepackage[pagebackref,breaklinks,colorlinks,allcolors=cvprblue]{hyperref}

\def\paperID{18131} 
\def\confName{CVPR}
\def\confYear{2025}

\usepackage{glossaries}
\usepackage{bbm}

\definecolor{lightgray}{gray}{0.9}
\definecolor{linecolor}{rgb}{0.82, 0.94, 0.75}

\newcommand{\pma}{{PointMamba}}

\hypersetup{
    colorlinks=true,
    linkcolor=red,
    filecolor=magenta,      
    urlcolor=magenta,
}

\newcommand{\CC}{\mathcal{C}}
\usepackage{multirow} 

\newacronym{ssl}{SSL}{Self-Supervised Learning}
\newacronym{sota}{SOTA}{State-Of-The-Art}
\newacronym{ssm}{SSMs}{State Space Models}
\newacronym{nlp}{NLP}{Natural Language Processing}
\newacronym{mae}{MAE}{Masked Autoencoder}
\newacronym{ae}{AE}{Autoencoder}
\newacronym{s4}{S4}{Structured State Space Sequence Model}
\newacronym{ssr}{SSR}{Spectral Spatial Reordering}
\newacronym{hse}{HSE}{Hierarchical Segmentation Enhancement}
\newacronym{vr}{VR}{Virtual Reality}
\newacronym{ar}{AR}{Augmented Reality}
\newacronym{csm}{CSM}{Cross-Scan Module}
\newacronym{mim}{MIM}{Masked Image Modeling}
\newacronym{svm}{SVM}{Support Vector Machine}
\newacronym{sast}{SAST}{Surface-Aware Spectral Traversing}
\newacronym{tar}{TAR}{Traverse-Aware Repositioning}
\newacronym{fps}{FPS}{Farthest Point Sampling}
\newacronym{knn}{KNN}{K-Nearest Neighbours}
\newacronym{hlt}{HLT}{Hierarchical Local Traversing}
\newacronym{sst}{SST}{Spectral Spatial Traversing}

\usepackage{graphicx}
\newcommand*\rot{\rotatebox{90}}
\newcommand{\rottiny}[1]{\,\rotatebox{90}{\tiny #1}}

\newcommand{\mypar}[1]{\vspace{2.5pt}
\noindent\textbf{#1~}}

\newcommand{\Real}{\mathbb{R}}
\newcommand{\Loss}{\mathcal{L}}

\newcommand{\mr}[1]{\mathit{#1}}

\title{Spectral Informed Mamba for Robust Point Cloud Processing}

\author{
Ali Bahri \thanks{Correspondence to \href{mailto:ali.bahri.1@ens.etsmtl.ca}{ali.bahri.1@ens.etsmtl.ca}} \and 
Moslem Yazdanpanah \and
Mehrdad Noori \and  
Sahar Dastani \and 
Milad Cheraghalikhani \and 
Gustavo Adolfo Vargas Hakim \and  
David Osowiechi \and  
Farzad Beizaee \and
Ismail Ben Ayed \and 
Christian Desrosiers \and
LIVIA, ÉTS Montréal, Canada\\
International Laboratory on Learning Systems (ILLS)\\
}

\begin{document}
\maketitle
\begin{abstract}
\gls*{ssm} have shown significant promise in Natural Language Processing (NLP) and, more recently, computer vision. This paper introduces a new methodology that leverages Mamba and \gls{mae} networks for point-cloud data in both supervised and self-supervised learning. We propose three key contributions to enhance Mamba's capability in processing complex point-cloud structures. First, we exploit the spectrum of a graph Laplacian to capture patch connectivity, defining an isometry-invariant traversal order that is robust to viewpoints and captures shape manifolds better than traditional 3D grid-based traversals. Second, we adapt segmentation via a recursive patch partitioning strategy informed by Laplacian spectral components, allowing finer integration and segment analysis. Third, we address token placement in \gls{mae} for Mamba by restoring tokens to their original positions, which preserves essential order and improves learning. Extensive experiments demonstrate the improvements that our approach brings over state-of-the-art baselines in classification, segmentation, and few-shot tasks. The implementation is available at: \url{https://github.com/AliBahri94/SI-Mamba.git}.

\end{abstract}    
\section{Introduction}
\label{sec:intro}

The analysis of 3D point cloud data is fundamental for various applications, including autonomous driving \cite{qi2021offboard, shi2019pointrcnn}, VR/AR \cite{guo2020deep}, and robotics \cite{rusu20113d}. Compared to the organized structure of 2D images, point clouds consist of 3D coordinates without direct adjacency information forming an unordered bag. In recent years, considerable efforts have been dedicated to adapt deep learning models such as convolutional neural networks (CNNs) and Transformers to this type of data \cite{qi2017pointnet++,yu2022point,pang2022masked,zhang2022point,bahri2024geomask3d, bahri2024test}. Due to their permutation-invariant self-attention mechanism, Transformer networks are particularly well suited for the unordered nature of point clouds. However, the quadratic complexity of this mechanism, requiring the computation of a weight between each pair of tokens, impedes the application of these networks to large-sized inputs (e.g., 2D images or 3D point clouds represented by many patches). This has prompted researchers to explore more efficient solutions, including the Set Transformer \cite{lee2019set}, Sparse Transformer \cite{child2019generating}, Longformer \cite{beltagy2020longformer} and Sinkhorn Transformer \cite{tay2020sparse}. 

\begin{figure}[t]
	\centering
		\includegraphics[width=1.\linewidth]{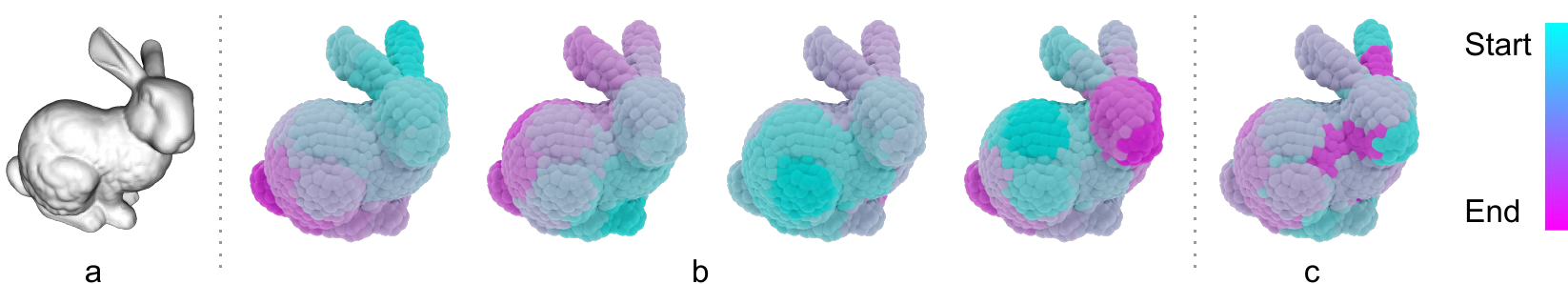}
	\caption{(a) \acrfull*{sast} applied to the patched point clouds of a mesh surface. (b) Traversal from left to right, based on the first to fourth non-constant smallest eigenvectors. (c) Traversal based on the largest eigenvector, forming a non-continuous sequence of tokens.}
	\label{fig:banner}
 \vspace*{-5mm}
\end{figure}

Recently, methods based on Structured State Space Sequence (S4) \cite{gu2021efficiently} such as Mamba \cite{gu2023Mamba} have gained significant traction as a more efficient alternative to Transformers  \cite{liu2024vMamba,zhu2024vision}.
So far, very few studies have investigated the potential of S4 approaches such as Mamba for 3D point clouds. Existing methods like Point-Mamba 
\cite{liang2024pointMamba} and PCM \cite{zhang2024point} extend the 2D grid-based traversal used for images to a 3D grid. However, this straightforward adaptation to point clouds suffers from three crucial problems.
 \textbf{First:} whereas patches from 2D images have adjacency information, which could be exploited by the grid-based traversal, the 3D point patches in point clouds offer a sparse representation of the object's surface, and nearby patches on a 3D grid are not necessarily adjacent on this surface. \textbf{Second:} in the absence of self-attention, task-specific performance is highly influenced by the nature of the token traversal strategy. For example, a traversal suitable for point cloud classification may not be effective for a local task such as point-level classification (i.e., segmentation). \textbf{Third:} due to the ``direction-sensitive'' nature of Mamba, the self-supervised \gls*{mae} pre-training step of leading point cloud models like Point-MAE \cite{pang2022masked} and Point-M2AE \cite{zhang2022point} cannot be used directly as there is no attention mechanism to learn the masked tokens' positions. 

The contribution of our work focuses on addressing these problems as follows:
\begin{enumerate}[itemsep=1pt,topsep=0pt]
    \item We introduce a \gls*{sast} strategy based on the Laplacian spectrum of a patch-connectivity graph. Compared to the 3D grid traversal of current approaches like Point-Mamba, our strategy is invariant to isometric transformations (e.g., choice of viewpoint) and better captures the object's surface manifold.
    \item We also present a \gls*{hlt} for point-level classification (segmentation) that partitions patches recursively based on their spectral coordinates. Unlike our \gls*{sast} strategy for classification, which considers Laplacian eigenvectors separately in different traversals, this \gls*{hlt} combines them in a single ordering for a more precise modeling of geometry.
    \item During the \gls*{mae}-based \gls*{ssl}, we propose a \gls*{tar} strategy to align the masked tokens according to their spectral adjacency. This strategy addresses the critical issue of spatial adjacency preservation unique to Mamba networks.
\end{enumerate}

\section{Related Works}
\label{sec:Related}

\begin{figure*}[!t]
	\centering
		\includegraphics[width=0.82\linewidth]{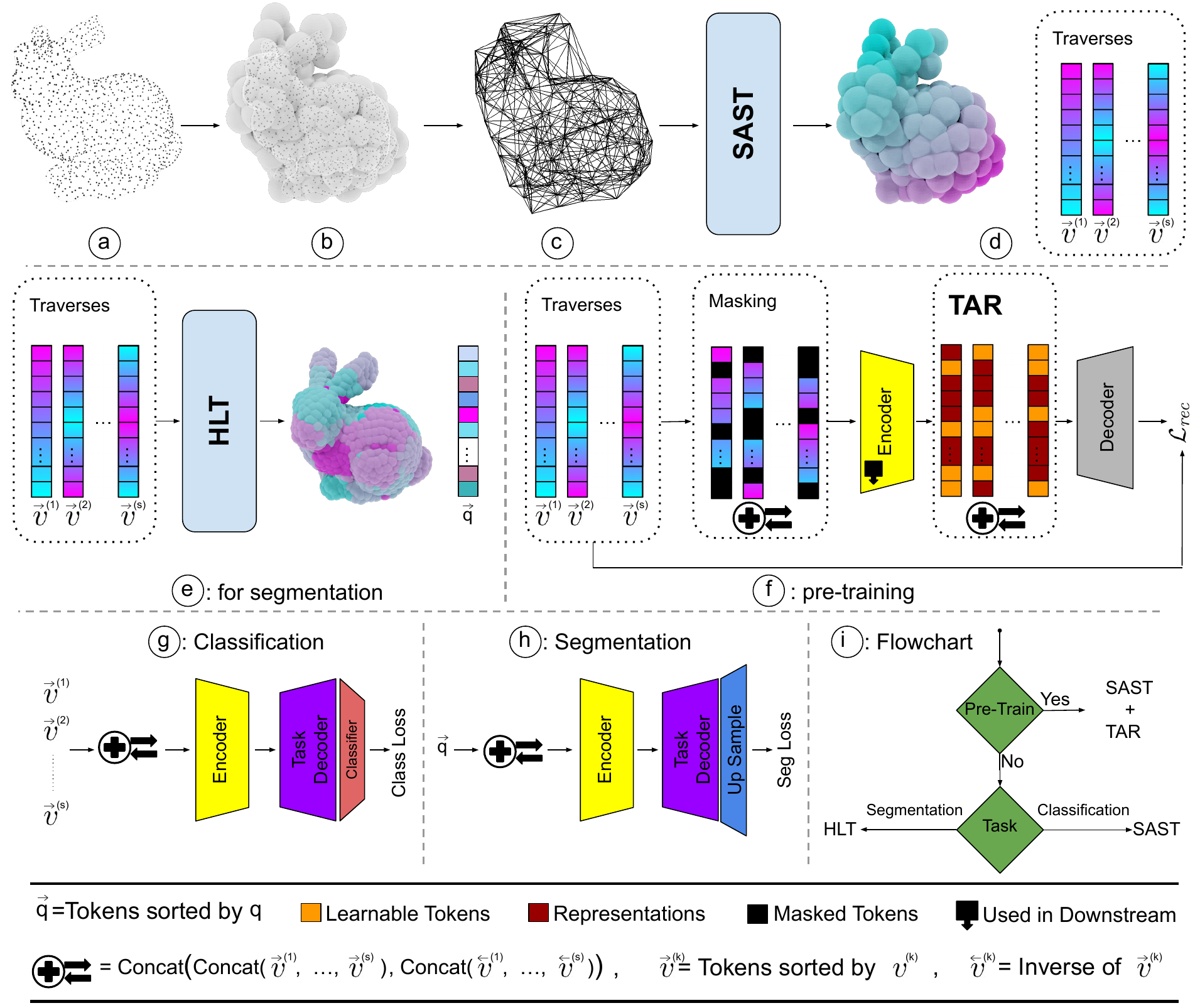}
	\caption{Overview of the proposed \acrfull{sst} method. (a) Point cloud, (b) Patchification, (c) Forming the adjacency graph, (d) Traversal based on \gls*{sast} using $s$ non-constant smallest eigenvectors, (e) \gls*{hlt} for segmentation tasks, (f) \gls*{tar} strategy for Masked Autoencoders. The process includes reverse and concatenation operations, with learnable tokens, representations, and masked tokens highlighted. (g) The classification task involves sorting tokens by different eigenvectors, concatenating them, and then feeding them into the network. (h) The segmentation task where HLT is applied on the tokens (\(\vec{q}\)) and \(\vec{q}\) is fed into the network. (i) A flowchart visualizing the techniques used in self-supervised learning and various downstream tasks.}
	\label{fig:method}
 \vspace*{-5pt}
\end{figure*}

\mypar{Deep Point Cloud Learning.}
Deep neural networks have increasingly been applied to point clouds, with early architectures 
like PointNet~\cite{qi2017pointnet} and PointNet++~\cite{qi2017pointnet++} inspiring further 
work~\cite{atzmon2018point, deng2023pointvector, landrieu2018large, li2018pointcnn, zhao2019pointweb} 
focused on capturing local context. Transformer-based frameworks~\cite{vaswani2017attention}, 
including the Point Transformer (v1--v3)~\cite{wu2023point, wu2022point, zhao2021point} 
and Stratified Transformer~\cite{lai2022stratified}, have set new benchmarks by integrating 
both local and global features. 
Recently, self-supervised pre-training has gained traction by leveraging masked point modeling~\cite{liu2023regress, tang2023point} to train Transformers on unlabeled data.

Building on the effectiveness of \gls*{mae} in Text and Image domains, Point-BERT \cite{yu2022point} presented a revolutionary method inspired by BERT \cite{devlin2018bert}, tailoring Transformers to 3D point cloud processing. 
Point-MAE \cite{pang2022masked} applied \gls*{mae}-style pre-training to 3D point clouds using a custom Transformer-based \gls*{ae} designed to reconstruct masked irregular patches.  The use of multi-scale masking and local spatial self-attention mechanisms in Point-M2AE \cite{zhang2022point} has led to \gls*{sota} results in 3D representation learning. Furthermore, I2P-MAE \cite{zhang2023learning} improved self-supervised point cloud processing with a masking strategy leveraging pre-trained 2D models through an Image-to-Point transformation. Point-GPT~\cite{chen2024pointgpt} introduced an auto-regressive generative pretraining (GPT) approach to address the unordered nature and low information density of point clouds. ACT~\cite{dong2022autoencoders} proposed a cross-modal knowledge transfer method using pretrained 2D or natural language Transformers as teachers for 3D representation learning. Finally, GeoMask3D \cite{bahri2024geomask3d} is a self-supervised point cloud method that uses a teacher-student model to select harder, geometrically intricate regions for masking, thereby improving feature representations.

\subsection{State Space Models}
\mypar{State Space Models.}
State-space models (SSMs) have been foundational in control theory and signal processing for modeling dynamic systems. Recently, they have been adapted to deep learning, with advancements like the Linear State-Space Layer (LSSL), which uses a continuous-time memorization framework based on the HiPPO operator \cite{gu2020hippo} to model long-range dependencies. However, LSSL’s high computational demands limit its practicality. To address this, the S4 model \cite{gu2021efficiently} introduced parameter normalization techniques, paving the way for structured SSMs with enhancements like complex-diagonal structures \cite{gupta2022diagonal, gu2022parameterization}, support for multiple-input/output \cite{gu2022parameterization}, and low-rank decomposition \cite{hasani2022liquid}. Recent developments include Mamba \cite{gu2023Mamba}, which achieves linear-time inference and efficient training through selection mechanisms and hardware-aware algorithms. MoE-Mamba further boosts efficiency by integrating a Mixture of Experts (MoE), outperforming both Mamba and Transformer-MoE models \cite{pioro2024moe}.

\mypar{SSMs for Vision Tasks}
The above-mentioned works primarily focused on the application of \gls*{ssm} to long-range or causal data types such as language and speech. In the field of vision, a notable study \cite{liu2024vMamba} proposed the VMamba model which features a Cross-Scan Module (CSM) for enhanced 1D selective scanning in 2D spaces and architectural optimizations that significantly improve its performance and speed across various visual tasks. Another significant paper is Vision Mamba \cite{zhu2024vision} which introduces a novel vision backbone called Vim utilizing bidirectional Mamba blocks. 


For point cloud analysis based on Mamba, two key works are Point-Mamba \cite{liang2024pointMamba} and PCM \cite{zhang2024point}. PointMamba introduces a simple approach to token reordering for point cloud analysis by strategically organizing point tokens based on a 3D grid. Similarly, PCM enhances Mamba with a Consistent Traverse Serialization (CTS) technique that converts 3D point clouds into 1D point sequences while maintaining spatial adjacency. 
Building upon these approaches, our method introduces the \acrfull{sst} strategy, which improves token ordering and maintains spatial adjacency within Mamba networks.
\section{Method}
\vspace{-5 pt}
We begin by outlining the fundamental concepts of \gls*{ssm} and spectral graph analysis which are at the core of our work. We then give an in-depth presentation of our \acrfull*{sast} strategy for point cloud processing that improves the model's robustness to isometric transformations and better captures the underlying manifold of the point cloud. 
Thereafter, we provide detailed specifications of our \acrfull*{hlt} strategy for point-level classification, which defines a more structured patch traversal order based on the recursive partitioning of spectral information. Finally, we introduce our \acrfull*{tar}, which improves the handling of learnable tokens in masked autoencoders within Mamba networks. ~\cref{fig:method} illustrates the overview of the proposed \acrfull{sst} method.
\subsection{Preliminaries}
\label{sec:Preliminaries}
\mypar{State Space Models (SSMs)} use a series of first-order differential equations to describe how the state of the linear, time-invariant system evolves over time:
\begin{equation}
\dot{h}(t)  = A h(t) + B x(t),\quad y(t) \, = \, C \dot{h}(t) + D x(t),
\end{equation}
Here, $\dot{h}(t)$  denotes the time derivative of the state vector $h(t)$. The matrices $A$, $B$, $C$, and $D$ are the weighting parameters. 

Due to their reliance on continuous data streams $x(t)$, SSMs are not natively equipped to handle discrete inputs represented as $\{x_0, x_1, \ldots\}$. This necessitates the use of a discretized SSM version for practical applications:
\begin{equation}
h_k = \bar{A} h_{k-1} + \bar{B} x_k,\quad
y_k  = \bar{C} h_k + \bar{D} x_k.
\end{equation}
The state space model in its discrete version utilizes a recursive function to link each state $h_k$ to its preceding state, encapsulated by the matrices $\bar{A} \in \mathbb{R}^{N \times N}$, $\bar{B} \in \mathbb{R}^{N \times 1}$, and $\bar{C} \in \mathbb{R}^{N \times 1}$, which are tuned parameter matrices. While matrix $\bar{D} \in \mathbb{R}^{N \times 1}$ may be employed as a residual connection, we follow previous work and exclude it from our model. The transition from a continuous signal representation $x(t)$ to a discrete sequence involves sampling $x(t)$ at intervals defined by $\Delta$, setting each discrete input as $x_k = x(k\Delta)$. This adjustment to a discrete framework results in revised matrix definitions:
\begin{equation}\label{eq:lti}
\bar{A} = (I - \frac{\Delta}{2} A)^{-1}(I + \frac{\Delta}{2} A), \bar{B} = (I - \frac{\Delta}{2} A)^{-1} \Delta B, \bar{C} = C.
\end{equation}
However, the fixed dynamics of Linear Time-Invariant (LTI) models, exemplified by the constant parameters $A$, $B$, and $C$ in Eq. (\ref{eq:lti}), restrict their capacity to selectively retain or discard relevant information, thereby limiting their contextual awareness. To improve content-aware reasoning, we use Mamba's selection mechanism that manages the propagation and interaction of information across the sequence dimension \cite{gu2023Mamba}.

\begin{figure}[!t]
	\centering
\includegraphics[width=1.0 \linewidth]{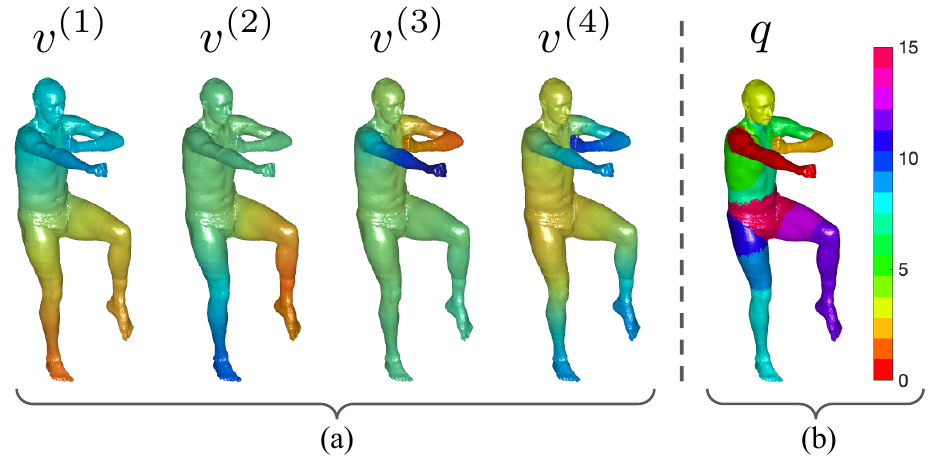}
	\caption{(a) Visualization of the four non-constant smallest Laplacian eigenvectors ($v^{(k)}$, $k=1,\ldots,4$) and (b) the discrete partitioning ($q$) of our HLT strategy combining the information of all four eigenvectors. Note: we assumed that patches contain a single point for better visualization.}
	\label{fig:eigenplot}
 \vspace*{-5mm}
\end{figure}

\vspace*{3pt}
\mypar{Spectral Graph Analysis.} Popularized by Chung in the 90s \cite{chung1997spectral}, spectral graph analysis characterizes the properties of a graph $G = (V,E)$ by the spectrum (eigenvalues and corresponding eigenvectors) of its Laplacian matrix $L$. This analysis can be understood as a discretized version of the Laplace-Beltrami Operator $\Delta$ of a function $f$ defined on a  Riemannian manifold:
\begin{equation}
\Delta f = \mathrm{div}(\mathrm{grad}  f) 
\end{equation}
where $\mathrm{grad} f$ is the gradient of $f$ and $\mathrm{div}$ the divergence on the manifold. The solution to the Laplacian eigenvalue problem $\Delta f =-\lambda f$, known as Helmholtz wave equation, is an eigenfunction corresponding to the natural vibration form of a homogeneous membrane with eigenvalue $\lambda$ \cite{reuter2005laplace}.

Following methods for spectral clustering \cite{ng2001spectral} and normalized cuts \cite{shi2000normalized}, we consider a weighted adjacency matrix $W: V\!\times\!V \to \Real_+$ where $W_{ij}=0$ if $(i,j)\not\in E$ to model the Euclidean distance of nearby patches (see Section \ref{sec:sast}). The Laplacian matrix of $G$ is defined as $L=D-W$ where $D$ is the diagonal \emph{degree} matrix such that $D_{ii} = \sum_j W_{ij}$. To account for variability in the scale of weights $W_{ij}$ or the distribution of node degrees $D_{ii}$, it is preferable to employ a normalized version of the Laplacian. In this work, we use the Random Walk Laplacian $L_{\mr{rw}} = I - D^{-1}W$ which has the following useful properties:
\begin{enumerate}
\item $L_{\mr{rw}}$ is positive semi-definite and has $|V|$ non-negative real-valued eigenvalues $0 =
\lambda_1 \leq \ldots \leq \lambda_{|V|}$;
\item 0 is an eigenvalue of $L_{\mr{rw}}$ with the constant vector as eigenvector and its multiplicity equals the number of connected components in the graph;
\item Following Courant's Nodal Line Theorem \cite{courant2008methods}, the $n$-th eigenmode of $L_{\mr{rw}}$ has at most $n$ poles of vibration;
\item The representation of a shape by the spectrum of $L_{\mr{rw}}$ is invariant to isometry (i.e., distance-preserving transformation).
\end{enumerate}

Our method uses the $s$ first non-constant eigenvectors of $L_{\mr{rw}}$ (i.e., the eigenvectors corresponding to the $s$ smallest non-zero eigenvalues) to define traversal orders for classification (Section \ref{sec:sast}) and segmentation (Section \ref{sec:hlt}) that are robust to the viewpoint (due to isometry invariance) and provide a smooth parametrization of the surface manifold. We consider the first eigenvectors as they encode low frequency information (by Courant's Nodal Line Theorem), making the resulting traversal more robust to shape variability and noise.   
~\cref{fig:banner} illustrates this concept: (a) shows the original mesh, (b) shows traversals based on the first to fourth non-constant smallest eigenvectors, and (c) shows traversal based on the largest eigenvector forming a non-continuous sequence of tokens.

\subsection{Point Cloud Patchification}
Given a point cloud $\mathcal{P} = \{p_i\}_{i=1}^{N_p}$, each point represented by 3D coordinates, we convert $\mathcal{P}$ to a reduced set of patches that can be processed more efficiently. Toward this goal, we employ the \gls*{fps} algorithm to select a subset $\mathcal{C} \subset \mathcal{P}$ of $N_c$ points offering a good coverage of the entire point cloud. These selected points will act as the centers of local patches within the point cloud. For each center point $p_{s_i} \in \CC$, we then identify $N_n$ nearest points $\mathcal{N}(p_{s_i}) \subset \mathcal{P}$ using the \gls*{knn} algorithm. Following this, each patch is defined as a center $p_{s_i}$ and its corresponding nearest-neighbors $\mathcal{N}(p_{s_i})$.

\subsection{\acrfull*{sast}}\label{sec:sast}
Current point cloud processing approaches using Mamba, such as Point-Mamba \cite{liang2024pointMamba} and PCM \cite{zhang2024point}, simply extend the 2D grid-based traversal for images to a 3D grid. As mentioned before, this naive strategy suffers from two issues: 1) the 3D grid is view dependent, thus rotating the point cloud or moving the camera yields a different traversal order; 2) unrelated patches may be adjacent in 3D space, hence can be traversed subsequently. To address these problems, we define a traversal order based on the Laplacian spectrum of the patch-connectivity graph.

In this graph, each node corresponds to a patch and the weighted adjacency matrix $W$ is defined using the Euclidean distance between patch centers. For patches $i$ and $j$ defined by center points $p_{s_i}$ and $p_{s_j}$, we add an edge $(i,j)$ if $p_{s_j}$ is among the $K$ nearest neighbors of $p_{s_i}$ or vice-versa. The weight of this edge is computed using a Gaussian kernel: $W_{ij} \, = \, \exp\big(\!-\!\|p_{s_i} - p_{s_j}\|^2_2/\sigma\big)$ where $\sigma$ is a hyperparameter controlling the kernel width.

Following Section \ref{sec:Preliminaries}, we compute the $s$ first non-constant eigenvectors of the Random Walk Laplacian $L_{\mr{rw}}$. This can be achieved efficiently using an iterative method like the Arnoldi algorithm \cite{golub2013matrix} by exploiting the following facts: 1) matrix $W$ is very sparse, and 2) only the first few eigenvectors need to be computed. Eigenvector $v^{(k)} \in \Real^{N_c}$, $k\in\{1,\ldots,s\}$, assigns an eigenfunction value $v^{(k)}_i$ to each patch $i$. In each Mamba block of our model, we perform two separate traversals of tokens (each token corresponds to an input patch) for \emph{every} eigenvector: a forward traversal by increasing value of $v^{(k)}_i$ and a reserve traversal by decreasing value of $v^{(k)}_i$. 
At the end of the block, we concatenate the tokens from the $s \times 2$ traversals.
The visual effect of each traversal is illustrated in~\cref{fig:banner}(b) and~\cref{fig:eigenplot}(a).


\label{Canonicalization}
\mypar{Canonicalization of spectrum.} Although the spectrum of $L_{\mr{rw}}$ forms an isometry-invariant representation of the surface manifold, this representation may be impacted by two sources of ambiguity: 1) the sign of eigenvectors is undetermined (i.e., if $v^{(k)}$ is an eigenvector, then so is $-v^{(k)}$), and 2) the order of eigenvectors with similar eigenvalues may vary. We address these two sources of ambiguity with the following canonicalization procedure. For the first one, we flip the sign of an eigenvector $v^{(k)}$ (i.e., $v^{(k)}:=-v^{(k)})$ if its first element is negative (i.e., $v^{(k)}_1 < 0$). To handle the second ambiguity, we first sort the eigenvectors by non-decreasing eigenvalue. We deal with eigenvalues having a mutliplicity greater than one by finding pairs of consecutive eigenvectors $v^{(k)}$, $v^{(k+1)}$ with near-identical eigenvalues (i.e., $|\lambda_k - \lambda_{k+1}| \leq \epsilon$). For such pairs, we flip the order if  $v^{(k)}_1 > v^{(k+1)}_1$. This reordering process is repeated until no further change occurs. 

\subsection{\acrfull*{hlt} for Segmentation}\label{sec:hlt}
While effective for classification tasks, the \gls*{sast} strategy considering each eigenvector in a \emph{separate} traversal may not capture the precise relationship between patches needed for segmentation. To address this issue, we introduce a Hierarchical Local Traversal (HLT) strategy that considers the full spectrum (all $s$ non-constant eigenvectors) simultaneously. 

Our strategy is inspired by the recursive binary partitioning technique of normalized cuts \cite{shi2000normalized}. Starting from the canonicalized spectrum (see Section \ref{Canonicalization}), tokens are first split based on the first eigenvector $v^{(1)}$, by comparing their corresponding value in $v^{(1)}$ with the mean value $\overline{v}^{(1)}$. This yields a binary partition of tokens $b^{(1)}_i\! =\mathbbm{1}\big(v^{(1)}_i \geq \overline{v}^{(1)}\big) \in \{0,1\}$ where $\mathbbm{1}$ is the indicator function. Each subgroup is then divided based on the mean of the second eigenvector $v^{(2)}$, and so on for other eigenvectors. This partitioning process can be seen as building a binary tree, where each level corresponds to a different eigenvector and leaf nodes $i$ are uniquely identified by the sequence of bits $b_i\! = \![b^{(1)}_i, \ldots, b^{(s)}_i]$ on the path from the root to the leaf. Our HLT method traverses groups of leaf nodes (groups of tokens) sequentially based on the lexicographic order of their binary code (e.g., $[0000]$, $[0001]$, $[0010]$, $[0011]$, $\ldots$ in the case of four eigenvectors). For convenience, we convert binary codes $b_i$ to a non-negative integer $q_i$ (e.g., $\mr{bin2Int}([0011])\!=\!3$) and define two traversal orders, by increasing or decreasing values of $q_i$. 

For $s$ eigenvectors, the HLT strategy described above divides tokens into $2^s$ segments which are traversed sequentially. In the best case scenario, $\lceil\log_2(N_c)\rceil$ eigenvectors are thus needed to split tokens into individual segments. However, it may happen that multiple tokens fall in the same segment, especially when using fewer eigenvectors. In such case, one can further sort tokens \emph{within} each segment, for example, using the values of the first eigenvector (i.e., $v^{(1)}$). In our implementation, we simply sort these tokens randomly to add stochasticity in the training. This section is illustrated in ~\cref{fig:method} (e), and ~\cref{fig:eigenplot} (b). 

As shown in ~\cref{fig:eigenplot}, the first Laplacian eigenvectors encode high-level spatial relations (e.g., bottom \emph{vs.} top, left \emph{vs.} right, torso \emph{vs.} limbs, etc.). In the SAST, because these eigenvectors are used in \emph
{separate} traversals, the network may not be able to differentiate specific regions/parts of the point cloud. In contrast, our HLT strategy can capture such specific parts (e.g., head, left or right arm/thigh/calf, pelvis, etc.).

\subsection{\acrfull*{tar} for Masked Autoencoders}
Following state-of-art Transformers for point cloud processing, such as Point-MAE \cite{pang2022masked} and Point-M2AE \cite{zhang2022point}, our method leverages self-supervised pretraining based on \gls*{mae} to boost performance. In Transformer-based approaches, the learnable tokens of masked patches can be inserted in any position of the sequence (typically at the end) as the self-attention mechanism can still attend to all tokens irrespective of their positions. However, this approach presents a significant problem in Mamba networks which are sensitive to the traversal order of tokens. We handle this problem via a \gls*{tar} strategy that improves the placement of learnable tokens in \gls*{mae} within Mamba networks. Specifically, we restore the learnable tokens to their original positions rather than appending them at the end of the sequence. This ensures that the essential order of tokens is maintained, preserving spatial adjacency and enhancing learning effectiveness within Mamba networks.

The proposed \gls*{tar} strategy selects an arbitrary traversal order and randomly masks a subset of $N_m$ tokens with the same masking ratio as the transformer-based \gls*{mae}s. These tokens are then removed from the sequence, and their positions are recorded. Afterwards, the remaining (visible) tokens are fed to the encoder that outputs their representation. Before reconstructing the point cloud using the decoder, we reinsert the learnable tokens in the sequence at their recorded position. The set of masked patches is used for other traversal orders. This procedure can be seen in ~\cref{fig:method} (f). Following previous work, we measure the reconstruction error for masked patches using the Chamfer distance:
\begin{equation}
    \Loss_{\mr{rec}} \, = \, \frac{1}{N_m} \sum_{i=1}^{N_m}  \mathrm{Chamfer}(\mathcal{S}_i, \,\hat{\mathcal{S}}_i),
\end{equation}
where $\mathcal{S}_i \in \Real^{N_n\times 3}$ is the set of points forming the $i$-th masked patch and $\hat{\mathcal{S}}_i$ the reconstructed output for these points. The Chamfer distance between two sets of points $\mathcal{S}$ and $\hat{\mathcal{S}}$ is defined as
\begin{align}
 \mathrm{Chamfer}(\mathcal{S},\mathcal{S}') \, = \, 
 &\sum_{p \in \mathcal{S}} \, \min_{p' \in \mathcal{S}'} \| p - p'\|_2^2 \nonumber \\
 &+ \, \sum_{p' \in \mathcal{S}'} \, \min_{p \in \mathcal{S}} \| p - p'\|_2^2.
\end{align}

\begin{table}[b!]
    \centering
    \caption{\textbf{Few-shot classification on ModelNet40}. We report the average accuracy (\%) and standard deviation (\%) of 10 independent experiments. {`$^*$'} denotes reproduced results. A\rottiny{std} represents the average (A) and standard deviation (std), respectively.}
    \setlength{\tabcolsep}{2.2mm}
    \resizebox{.88\columnwidth}{!}{
    \begin{tabular}{lcccc}
        \toprule
       \multirow[b]{2}{*}{\ Method}  & \multicolumn{2}{c}{5-way} & \multicolumn{2}{c}{10-way} \\
       \cmidrule(l{4pt}r{4pt}){2-3}\cmidrule(l{4pt}r{4pt}){4-5}
        & \rot{10-shot} & \rot{20-shot} & \rot{10-shot} & \rot{20-shot} \\
        \midrule
        \ DGCNN~\cite{wang2019dynamic} & 91.8\rottiny{3.7} & 93.4\rottiny{3.2} & 86.3\rottiny{6.2} & 90.9\rottiny{5.1} \\
        \addlinespace[0.3em]
        {\renewcommand{\arraystretch}{0.7} 
        \begin{tabular}[l]{@{}l@{}} DGCNN\,+\,OcCo~\cite{wang2021unsupervised}\end{tabular}} & 91.9\rottiny{3.3} & 93.9\rottiny{3.1} & 86.4\rottiny{5.4} & 91.3\rottiny{4.6} \\
        \midrule
        \addlinespace[0.3em]
        \ Transformer~\cite{yu2022point} & 87.8\rottiny{5.2} & 93.3\rottiny{4.3} & 84.6\rottiny{5.5} & 89.4\rottiny{6.3} \\
        \addlinespace[0.3em]
        {\renewcommand{\arraystretch}{0.7}
        \begin{tabular}[l]{@{}l@{}} Transf.\,+\,OcCo~\cite{yu2022point}\end{tabular}} & 94.0\rottiny{3.6} & 95.9\rottiny{2.3} & 89.4\rottiny{5.1} & 92.4\rottiny{4.6} \\
        \addlinespace[0.3em]
        \  Point-BERT~\cite{yu2022point} & 94.6\rottiny{3.1} & 96.3\rottiny{2.7} & 91.0\rottiny{5.4} & 92.7\rottiny{5.1} \\
        \  Point-M2AE ~\cite{zhang2022point} & 96.8\rottiny{1.8} & 98.3\rottiny{1.4} & 92.3\rottiny{4.5} & 95.0\rottiny{3.0} \\
        \  Point-MAE ~\cite{pang2022masked} & 96.3\rottiny{2.5} & 97.8\rottiny{1.8} & 92.6\rottiny{4.1} & 95.0\rottiny{3.0} \\
        \addlinespace[0.3em]
        \midrule
        \  \pma {$^*$} \cite{liang2024pointMamba} & 95.9\rottiny{2.1} & 97.3\rottiny{1.9} & 91.6\rottiny{5.3} & 94.5\rottiny{3.5} \\
        \rowcolor{gray!15}
        \  Ours & \textbf{96.4\rottiny{2.7}} & \textbf{98.5\rottiny{1.5}} & \textbf{92.0\rottiny{5.1}} & \textbf{95.1\rottiny{3.6}} \\
        \bottomrule
    \end{tabular}
    }
    \label{tab:fewshot}
\end{table}

\begin{table}[b!]
\centering
  \setlength{\tabcolsep}{2.2mm}
  \caption{\textbf{Part segmentation on the ShapeNetPart~\cite{yi2016scalable}}. The mIoU for all instances (Inst.) is reported. HLT is Hierarchical Local Traversing for segmentation.}

      \label{tab:segmentation}
      \resizebox{.88\columnwidth}{!}{
    \begin{tabular}{lccc}
    \toprule
     Methods  & {Backbone}  & {FLOPs (G)} & {mIoU} \\
    \midrule\midrule
     \multicolumn{4}{c}{\textit{Training from scratch}} \\
 \midrule
     PointNet~\cite{qi2017pointnet} & -  &-  & 83.7 \\
     PointNet++~\cite{qi2017pointnet++}  & -  &- & 85.1\\
     DGCNN~\cite{wang2019dynamic} & - & - & 85.2\\
     APES~\cite{wu2023attention} & - & - & 85.8\\
     Point-MAE~\cite{pang2022masked}& Transformer &15.5 & 85.7\\

  \midrule
 \pma~\cite{liang2024pointMamba} &Mamba & 14.3  & 85.8\\
\rowcolor{gray!15}
 \ Ours (HLT) &Mamba & 14.3  & \textbf{85.9}\\
 \midrule\midrule
  \multicolumn{4}{c}{\textit{Training from pretrained}} \\
 \midrule
 Transformer~\cite{yu2022point} & Transformer  & 15.5  & 85.1 \\
 Point-BERT~\cite{yu2022point} & Transformer  & 15.5 & 85.6\\
 Point-MAE~\cite{pang2022masked} & Transformer  &15.5 & 86.1\\ 
 Point-M2AE~\cite{zhang2022point} & Transformer  &15.5 & 86.5\\ 
 Point-GPT-S~\cite{chen2024pointgpt} & Transformer & - & 86.2 \\
 ACT~\cite{dong2022autoencoders} & Transformer & - & 86.2 \\
 I2P-MAE~\cite{zhang2023learning} & Transformer & - & 86.8 \\
  \midrule
 \pma~\cite{liang2024pointMamba} &Mamba &14.3  &  86.0\\
 Ours (SAST)  &Mamba & 14.3 & 85.7\\
\rowcolor{gray!15}
 Ours (HLT)  &Mamba & 14.3 & \textbf{86.1}\\
    \bottomrule
    \end{tabular}
    }
    \vspace*{-7pt}
\end{table}

\section{Experiments}
\label{others}




\begin{table*}[ht]
  \centering
\footnotesize	
\caption{\textbf{Object classification on ScanObjectNN~\cite{uy2019revisiting}}. Accuracy (\%) is reported. \textsuperscript{$\dagger$} indicates that this method was fine-tuned without rotation augmentation.}
\setlength{\tabcolsep}{1.2mm} 
\label{tab:scanobjnn}
\resizebox{.88\textwidth}{!}{ 
\begin{tabular}{lccccccc}
\toprule
Methods & Backbone & Param. (M)& FLOPs (G) & OBJ-BG & OBJ-ONLY & PB-T50-RS & ModelNet40\\
\midrule
\multicolumn{8}{c}{\textit{Training from scratch}}\\
 \midrule
PointNet~\cite{qi2017pointnet} & - & 3.5 & 0.5  & 73.3  & 79.2  & 68.0 & 89.2\\
PointNet++~\cite{qi2017pointnet++} & - & 1.5 & 1.7 & 82.3  & 84.3 & 77.9 & 90.7\\
DGCNN~\cite{wang2019dynamic} & - & 1.8 & 2.4 & 82.8  & 86.2  & 78.1 & 92.9\\
PointNeXt~\cite{qian2022pointnext}  & -  & 1.4 & 1.6 & -     & -  & 87.7 & 92.9\\
PointMLP~\cite{ma2022rethinking}  & - &  13.2 & 31.4  & -    & -     & 85.4 & -\\
ADS~\cite{hong2023attention} & - & -  & -  &  - & -   & 87.5 & -\\
Transformer~\cite{yu2022point} &Transformer & 22.1 & 4.8 & 79.86 & 80.55 & 77.24 & -\\
Point-MAE~\cite{pang2022masked} & Transformer& 22.1 & 4.8 & 86.75 & 86.92 & 80.78 & 92.3\\
  \midrule
\pma~\cite{liang2024pointMamba} &Mamba & 12.3& 3.6 & 90.87 & 90.18 &  85.60 & 92.4\\
\rowcolor{gray!15}
Ours &Mamba & 12.3 & 3.6 & \textbf{92.25} & \textbf{91.39} & \textbf{87.30} & \textbf{92.7}\\
 \midrule
 \multicolumn{8}{c}{\textit{Training from pretrained}}\\
 \midrule
Transformer~\cite{yu2022point} &Transformer & 22.1 & 4.8 & 79.86 & 80.55 & 77.24 & 92.1 \\
Transformer-OcCo~\cite{yu2022point}~ &Transformer & - & - & 84.85 & 85.54 & 78.79 & -\\
Point-BERT~\cite{yu2022point} &Transformer &22.1 & 4.8 & 87.43 & 88.12 & 83.07 & 92.7\\
Point-M2AE\textsuperscript{$\dagger$}~\cite{zhang2022point} & Transformer& - & - & 91.22 & 88.81 &  86.43 & 94.0 \\
Point-MAE ~\cite{pang2022masked} &Transformer & 22.1 & 4.8 & 92.77& 91.22 & 89.04 & 93.2\\
  \midrule
\pma~\cite{liang2024pointMamba} &Mamba & 12.3 & 3.6 & 93.29 & 91.56 & 88.17 & 92.8\\
PCM \cite{zhang2024point} &Mamba & 12.3& 3.6 & - & - &  86.9 & -\\
\rowcolor{gray!15}
Ours &Mamba & 12.3 & 3.6 & \textbf{94.32} & \textbf{91.91} & \textbf{89.10} & \textbf{93.4}\\
  \bottomrule
\end{tabular}
}
\end{table*}

Several experiments are conducted to evaluate the proposed method. First, we pretrain the Point-Mamba network using our techniques on the ShapeNet~\cite{chang2015shapenet} training dataset. We then assess the performance of these pretrained models across a variety of standard benchmarks, including object classification, few-shot learning, and segmentation. Additionally, we train the model from scratch on downstream datasets to demonstrate the robustness and versatility of our method.
To have a fair comparison, we adopt the masking ratio (60\%) that was used in the Point-Mamba model. Moreover, a comprehensive analysis of the computational efficiency, runtime, and memory usage of our SAST approach is provided in the Supplementary Material.

\subsection{Pretraining Setup}
Following Point-Mamba, we adopt the ShapeNet~\cite{chang2015shapenet} dataset for the pretraining and assess the quality of the 3D representations produced by our approach through a linear evaluation on the ModelNet40 \cite{wu20153d} dataset.  
The linear evaluation is performed by a \gls*{svm} fitted on these features. This classification performance is quantified by the  accuracy metric. Augmentation in pretraining is \textit{scale and transform}, while in finetuning it is \textit{rotation}.

\subsection{Downstream Tasks}
\mypar{Object Classification on Real-World Dataset.} To evaluate our method for point clouds, we test it on the ScanObjectNN dataset \cite{uy-scanobjectnn-iccv19}, as described in previous studies. 
The augmentation used during training is random rotation. The results, presented in ~\cref{tab:scanobjnn}, show that our strategy significantly improves object classification accuracy in both training from scratch and fine-tuning scenarios. These findings highlight the effectiveness of our approach in enhancing the model's ability to identify and classify objects across various backgrounds, demonstrating its robustness in complex real-world scenarios. 

\vspace*{3pt}
\mypar{Object Classification on Clean Objects Dataset.} 
We also evaluated our method on the ModelNet40 \cite{wu20153d} dataset, following the protocols established in previous works. The augmentation used during training is scale and transform. As shown in \cref{tab:scanobjnn}, our approach achieves notable enhancements on this challenging dataset compared to both the original Point-Mamba and the Transformer-based Point-MAE. This demonstrates the robustness and effectiveness of our method when applied to the Point-Mamba network.

\vspace*{3pt}
\mypar{Few-shot Learning.} We conducted few-shot learning experiments on ModelNet40 \cite{wu20153d} dataset, adhering to the protocols of previous studies \cite{pang2022masked, zhang2022point, liu2022masked}. 
The results of our few-shot learning experiments are presented in ~\cref{tab:fewshot}. Despite the competitive nature of this benchmark, our method demonstrated outstanding performance across all tested scenarios. As shown in ~\cref{tab:fewshot}, our Mamba-based method achieves results comparable to or exceeding those of transformer-based methods (Point-MAE and Point-M2AE).

\begin{figure*}[t!] 
    \begin{minipage}{0.5\textwidth} 
        \centering
        \includegraphics[width=.8\linewidth]{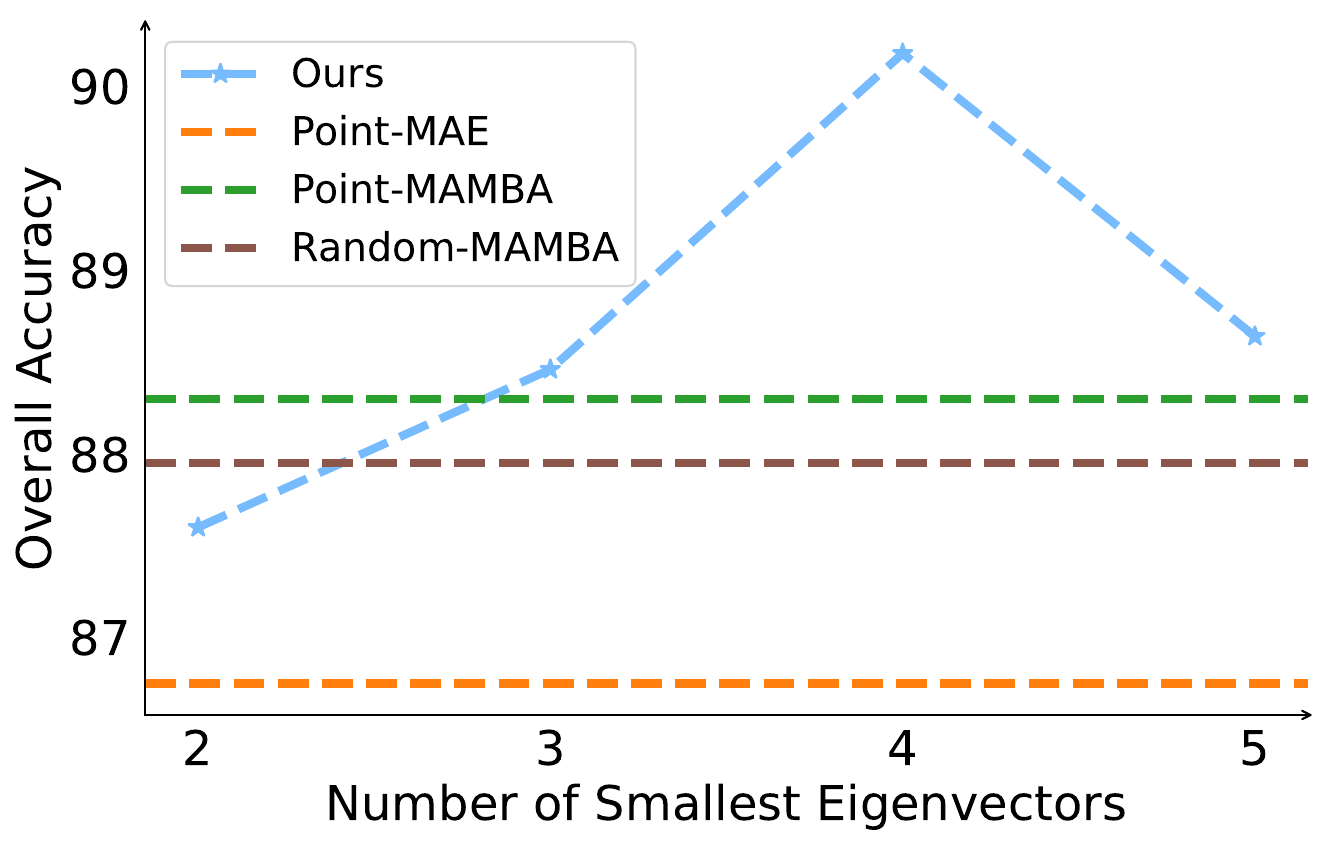}
    \end{minipage}%
    \begin{minipage}{0.5\textwidth} 
        \centering
        \includegraphics[width=.8\linewidth]{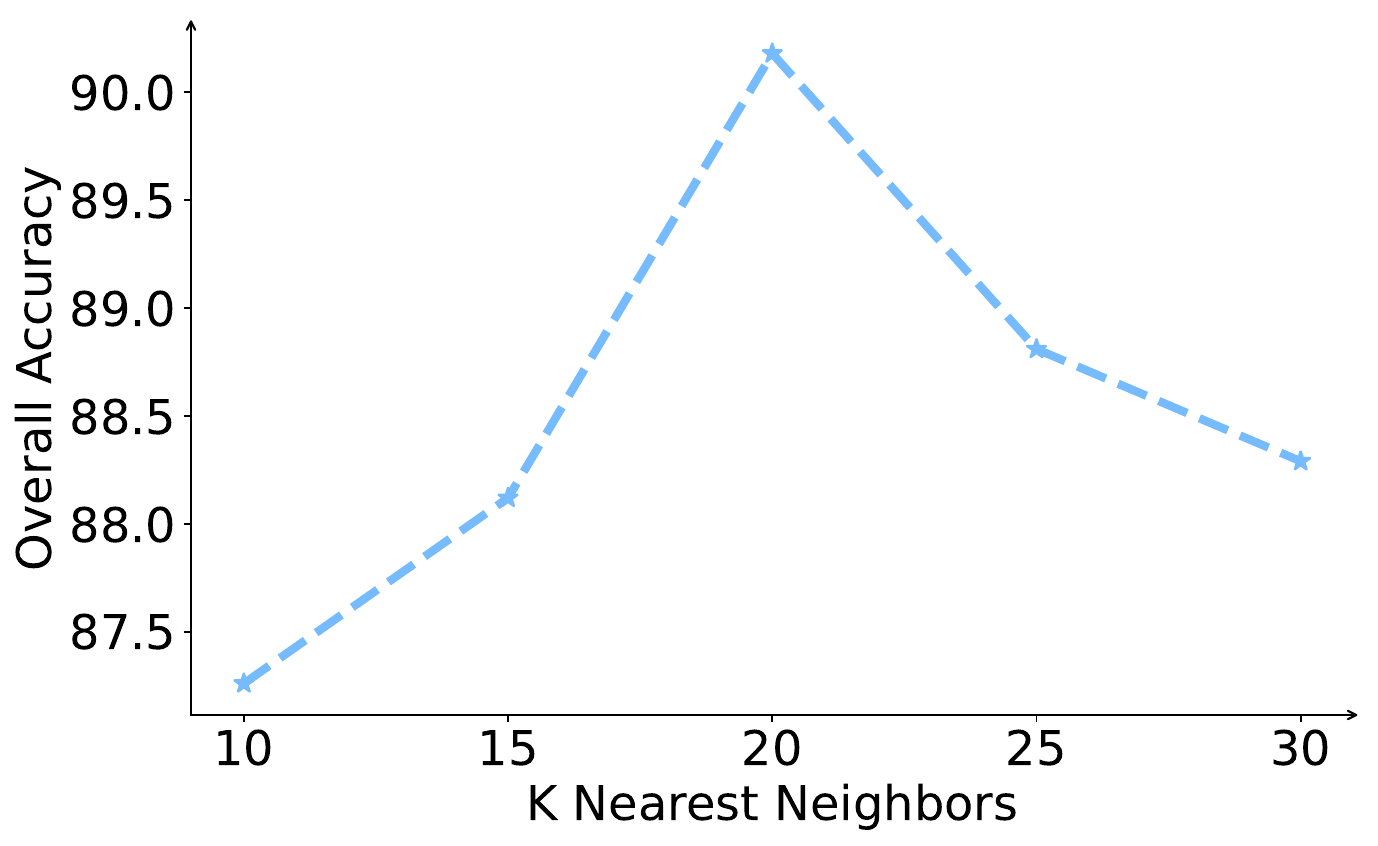}
    \end{minipage}
    \caption{Analysis of the number of non-constant smallest eigenvectors and comparison with previous methods (\textit{Left}) and analysis of the number of nearest neighbors $K$ (\textit{Right}).}
    \label{fig:ablation_1}
\end{figure*}

\vspace{-5pt}
\begin{figure*}[t!] 
    \begin{minipage}{0.5\textwidth} 
        \centering
        \includegraphics[width=0.8\linewidth]{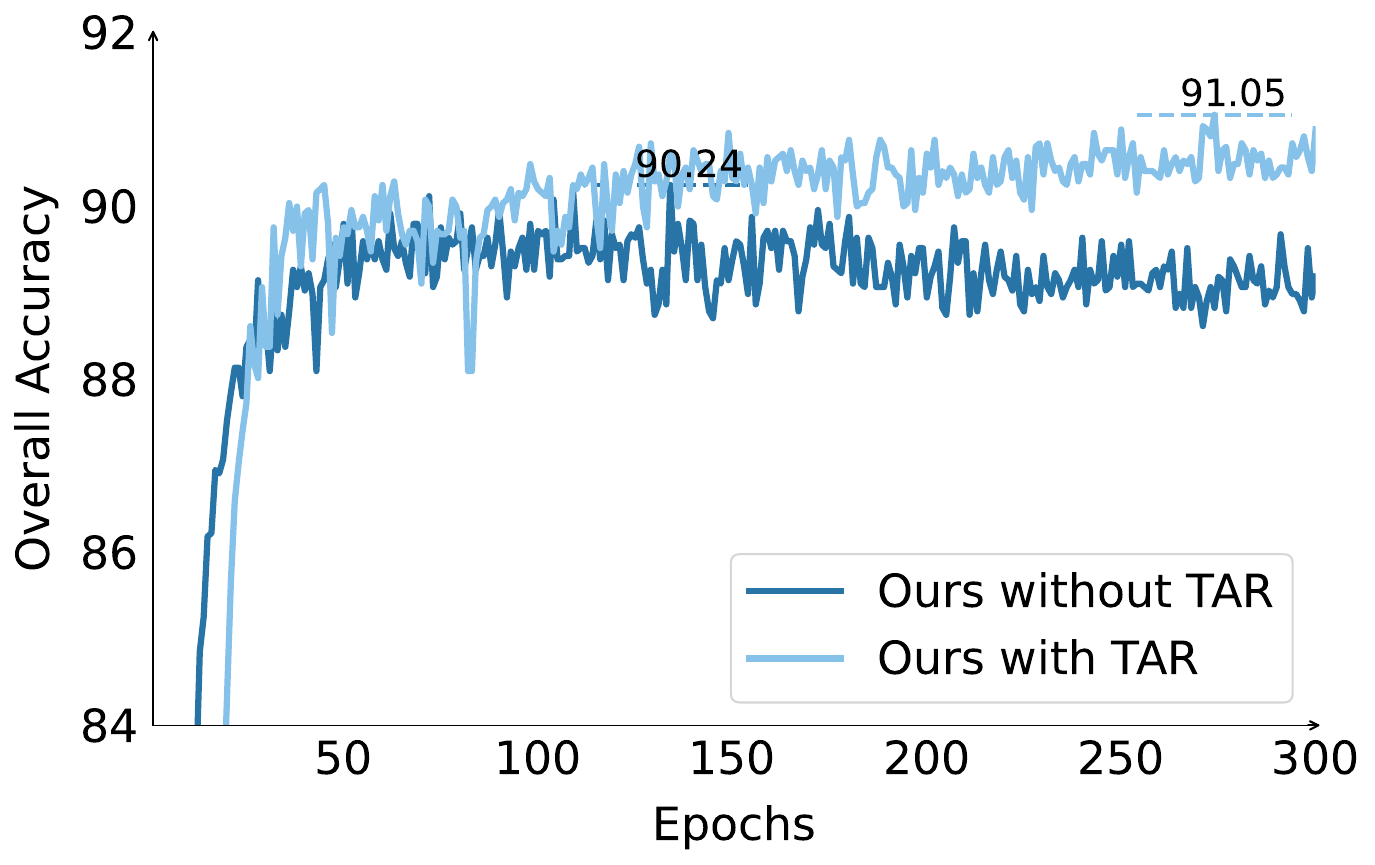}
    \end{minipage}%
    \begin{minipage}{0.5\textwidth} 
        \centering
        \includegraphics[width=0.8\linewidth]{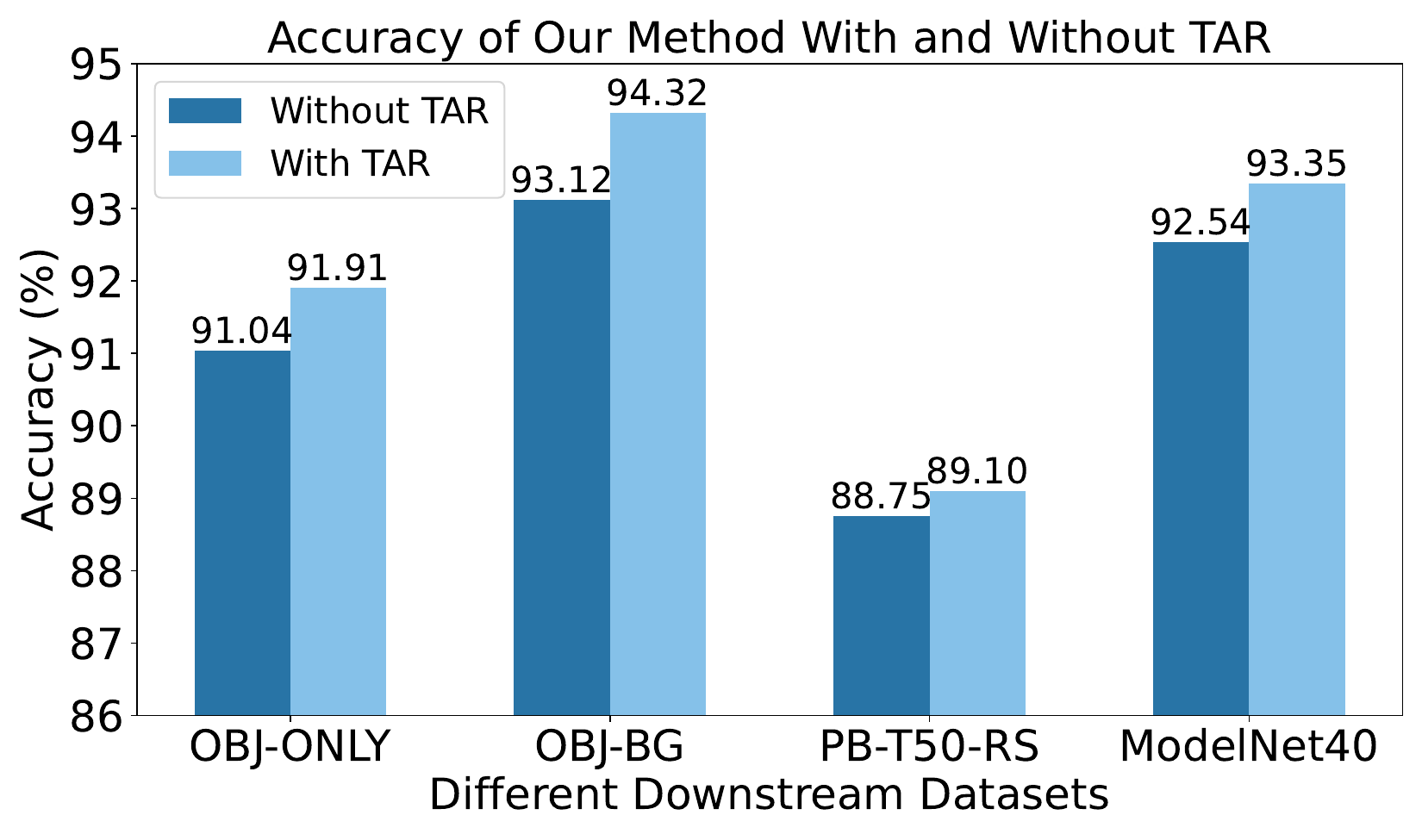}
    \end{minipage}
    \caption{The effect of the \gls*{tar} strategy in the pretraining phase (\textit{Left}) and in finetuning (\textit{Right}).}
    \label{fig:pretrain_ablation}
    \vspace{-12pt}
\end{figure*}

\vspace*{3pt}
\mypar{Part Segmentation.} We evaluated our method's representation learning ability on the ShapeNetPart dataset~\cite{yi2016scalable} using standard experimental settings~\cite{qi2017pointnet, qi2017pointnet++, yu2022point}. ~\cref{tab:segmentation} shows that performance gains in prior methods are minimal, highlighting the dataset's challenge. For instance, Point-GPT \cite{chen2024pointgpt} and ACT \cite{dong2022autoencoders}, despite being state-of-the-art and complex methods, show only marginal improvements over each other and other state-of-the-art approaches. 
Similarly, I2P-MAE \cite{zhang2023learning}, which combines 3D data with 2D information, shows limited improvement over Point-M2AE. 
For the ``Training from scratch'' setting, our method outperforms several state-of-the-art approaches. In the ``Training from pretrained'' setting, we further demonstrate the effectiveness of \gls*{hlt} strategy compared to \gls{sast} in the segmentation task.

\subsection{Ablation Studies}
In this section, we aim to investigate the effects of different parameters on our method. We will focus on two key aspects: the effect of the number of non-constant smallest eigenvectors and the adjacency matrix used in the \gls*{sast} strategy, and the analysis of the \gls*{tar} strategy. 

\vspace*{3pt}
\mypar{Analysis of Eigenvectors and Graph.} 
One of our ablation studies investigates the impact of the number of non-constant smallest eigenvectors used in the \gls*{sast} and \gls*{tar} strategies. As depicted in ~\cref{fig:ablation_1} (\textit{Left}), the best performance is achieved when using the four non-constant smallest eigenvectors (blue line) for traversing. When the number of non-constant smallest eigenvectors increases beyond four, the performance drops. This is because the additional eigenvectors are closer to the largest eigenvectors, which are less smooth and do not capture the most significant structural variations effectively.
The significantly lower performance of Point-MAE \cite{pang2022masked}, Point-Mamba \cite{liang2024pointMamba}, and Point-Mamba with random traversing (Random-Mamba) compared to ours underscores the importance of appropriate traversing for Mamba networks.

Another study examines the impact of the number of nearest neighbors used in creating the adjacency matrix. As shown in ~\cref{fig:ablation_1} (\textit{Right}), the best performance (blue line) is achieved with 20 nearest neighbors. The model's accuracy increases as the number of nearest neighbors increases from 10 to 20, reaching its peak at 20 nearest neighbors. Beyond this point, the performance starts to decline. All these experiments are trained and evaluated from scratch on the ScanObjectNN dataset~\cite{uy-scanobjectnn-iccv19} (OBJ-BG) with \textit{scale and transform} augmentation.

\vspace*{3pt}
\mypar{Analysis of \gls*{tar} Strategy.} 
One of the studies examines the impact of the \gls*{tar} strategy on the model's performance. As shown in ~\cref{fig:pretrain_ablation} (\textit{Left}), the accuracy of the model with and without the \gls*{tar} strategy is plotted against the number of epochs. The model incorporating the \gls*{tar} strategy (light blue line) demonstrates superior performance compared to the model without \gls*{tar} (dark blue line).
This figure relates to the pretraining phase on the ShapeNet ~\cite{chang2015shapenet} dataset, which is subsequently tested on the ModelNet \cite{wu20153d} dataset using a \gls*{svm}. 

Additionally, we evaluate the effect of the \gls*{tar} strategy by fine-tuning pretrained models (with or without \gls*{tar}) on four downstream tasks, each corresponding to a distinct dataset.
As shown in~\cref{fig:pretrain_ablation} (\textit{Right}), the model pretrained with \gls*{tar} (light blue line) consistently achieves higher accuracy than the model trained without \gls*{tar} (dark blue line).
This demonstrates that the model pretrained with the \gls*{tar} strategy has learned more meaningful features, leading to better performance in the downstream tasks.



\section{Conclusion and Limitation}
We introduced three strategies to enhance Mamba networks for point cloud data: isometry-invariant token traversal, recursive patch partitioning for segmentation, and improved learnable token placement, each contributing to a robust feature extraction framework. Our methods demonstrate superior performance over state-of-the-art baselines in classification, segmentation, and few-shot tasks. A limitation of our approach lies in the selection of the neighborhood size \( K \), which is treated as a hyperparameter. While tuning \( K \) per dataset helps optimize performance, it may limit generalizability across diverse data distributions and require careful adjustment to maintain robustness.

{
    \small
    \bibliographystyle{ieeenat_fullname}
    \bibliography{main}
}

\clearpage
\setcounter{page}{1}
\setcounter{section}{0}
\renewcommand{\thesection}{\Alph{section}}
\renewcommand{\thesubsection}{\thesection.\arabic{subsection}}
\maketitlesupplementary

\section{Computational Efficiency, Runtime, and Memory Usage}

In this section, we conducted a comprehensive analysis regarding the computational efficiency, runtime, and memory usage of our \gls{sast} approach. 
The focus of these experiments is to assess the overhead introduced by our \gls{sast} in comparison to the Point-Mamba backbone.

Our \gls{sast} strategy employs SciPy's sparse eigen-solver (function \texttt{eigs} implementing the implicitly restarted Arnoldi algorithm) to compute the first eigenvectors of the graph Laplacian. As highlighted in the main paper, this operation is not expensive since the size of this matrix depends on the number of patches which is much less than the original number of points (128 vs.\,2048). Additionally, even when increasing the number of patches (tokens), the computational overhead of this step is limited due to three reasons: \emph{i}) the Laplacian matrix is very sparse as we only consider the $K$ nearest neighbors of each patch ($K \approx 20$), \emph{ii}) we only compute the first $k$ eigenvectors ($k \approx 5$ in our experiments), and \emph{iii}) this computation is done \textbf{only once} for each point cloud in a \textbf{pre-processing step}. 

\begin{figure*}[h]
    \centering
    \includegraphics[width=.8\textwidth]{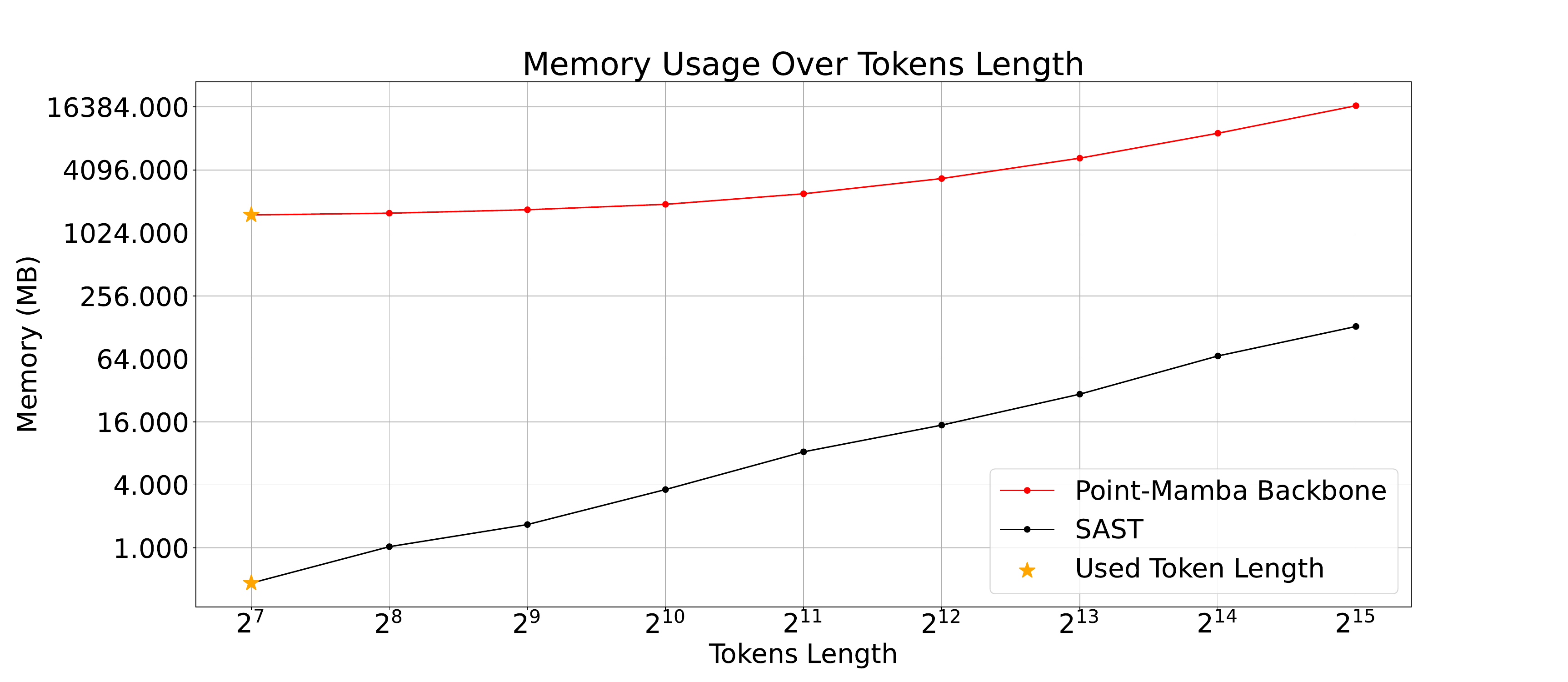}
    \caption{Memory Usage Over Tokens Length. Both axes are scaled by \(\log_2\) for better visualization.}
    \label{fig:memory_usage}
\end{figure*}

\mypar{Memory Usage:}
As shown in \cref{fig:memory_usage}, the memory usage of our \gls{sast} strategy (black line) is significantly lower than the memory usage of the Point-Mamba backbone (red line). When increasing the number of patches along the x-axis, our strategy based on a sparse eigen-solver does not require substantially more memory compared to the backbone. The star in this figure shows the used number of tokens in downstream tasks.

\begin{figure*}[!h]
    \centering
    \includegraphics[width=.8\textwidth]{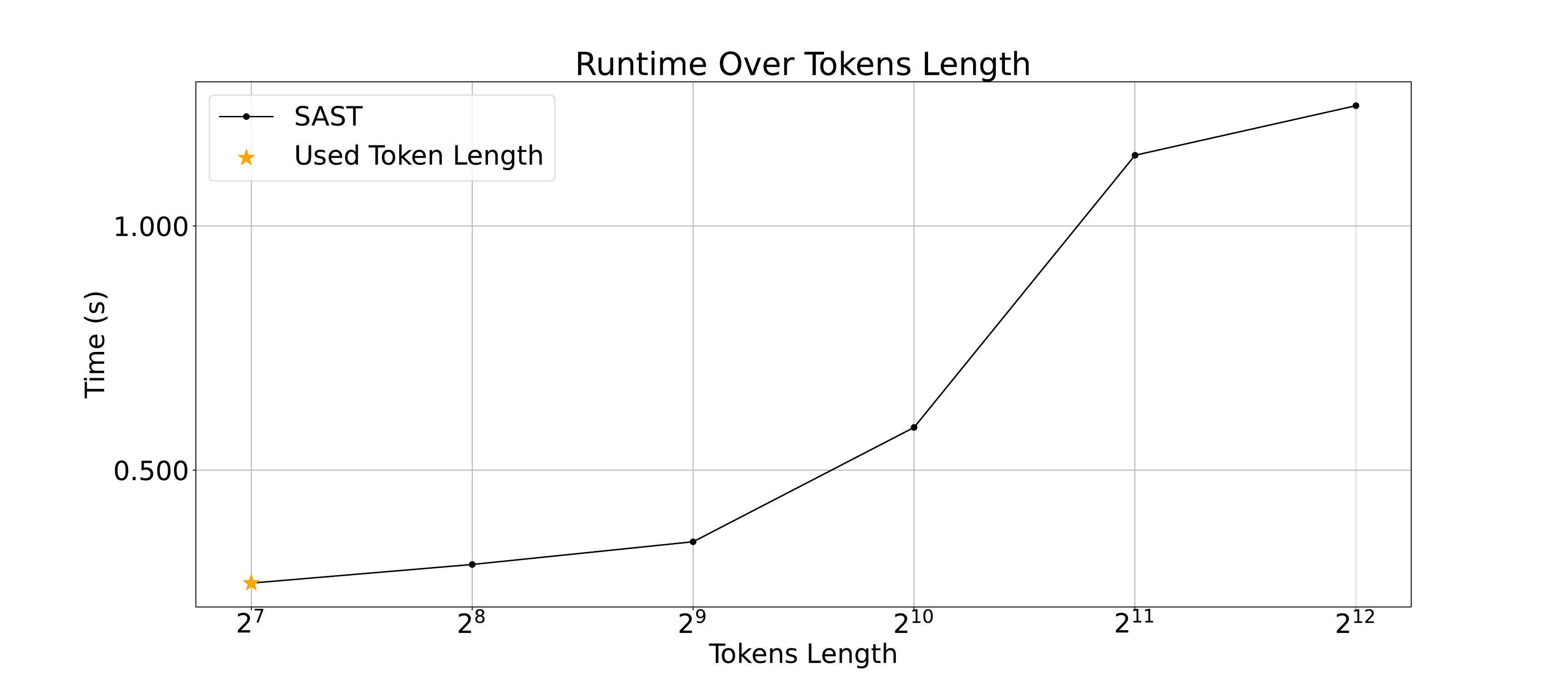}
    \caption{Runtime Over Tokens Length. Both axes are scaled by \(\log_2\) for better visualization. 
}
    \label{fig:runtime}
\end{figure*}

\mypar{Runtime:} Our \gls{sast} strategy, which can be implemented in the data loader and ran in parallel on CPU, is also fast. As can be seen in \cref{fig:runtime}, the runtime of \gls{sast} scales well when increasing the number of patches (tokens), and only a small amount of runtime is added in training or inference for the token length of 128 used in our main experiments (yellow star).

\begin{figure*}[!h]
    \centering
    \includegraphics[width=0.8\textwidth]{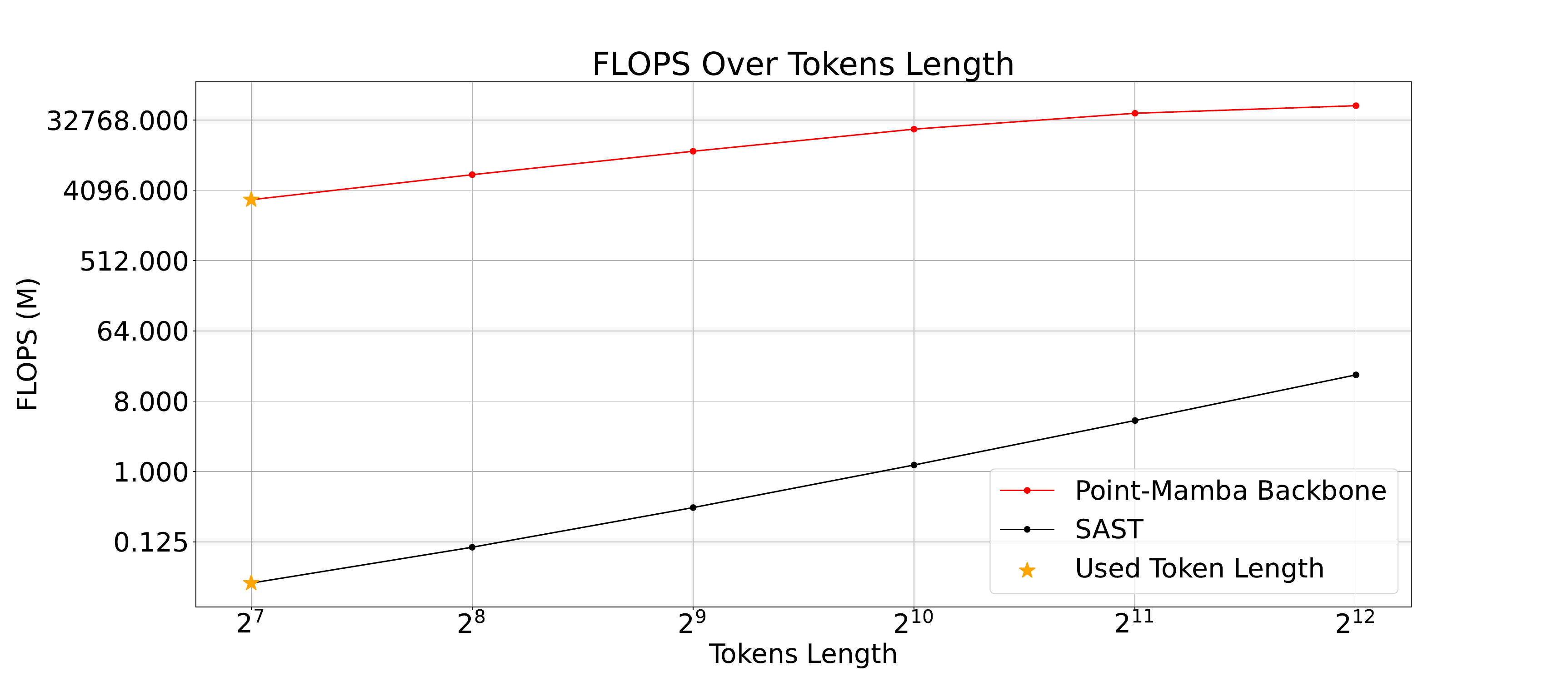}
    \caption{FLOPS Over Tokens Length. The horizontal axis is scaled by \(\log_2\) for better visualization. 
}
    \label{fig:flops}
\end{figure*}

\mypar{FLOPS:}
\cref{fig:flops} presents the relationship between FLOPS and token length for both the Point-Mamba backbone and our \gls{sast} method. 
Compared to running the Point-Mamba back, our \gls{sast} demonstrates a more gradual increase in FLOPS due to the use of sparse computations and the low number of eigenvectors involved. Once again, this shows the limited overhead of incorporating \gls{sast}, even as token length increases. 



\section{Additional Ablation Study}

\mypar{The Effect of HLT on Classification:} 
In this section, we investigate the effect of the \gls{hlt} strategy on the classification task. The results for \gls{hlt} on the ObjectNN dataset are shown in \cref{tab:scanobjnn_hlt}. As observed, the \gls{hlt} strategy underperforms compared to SAST across all three settings (OBJ-BG, OBJ-ONLY, and PB-T50-RS), regardless of whether the model is trained from scratch or pretrained. 

This performance gap highlights the limitations of the \gls{hlt} strategy in tasks requiring global understanding, such as classification. Specifically, \gls{hlt} processes high-level information from all eigenvectors simultaneously in a \textbf{single traversal order} (forward and backward), which is effective for segmentation tasks but may lead to insufficiently distinct feature representations for global classification. In contrast, SAST processes information from different eigenvectors in \textbf{separate traversals}, enabling better representation of high-level structures critical for classification tasks.

\begin{table}[htbp]
  \centering
\footnotesize	
\caption{Object classification on ScanObjectNN. Accuracy (\%) is reported. }
\setlength{\tabcolsep}{0.6mm}
\label{tab:scanobjnn_hlt}
\begin{tabular}{lcccccc}
\toprule
Methods & Backbone & OBJ-BG & & OBJ-ONLY & & PB-T50-RS \\
\midrule\midrule
\multicolumn{7}{c}{\textit{Training from scratch}}\\
 \midrule

\ Ours (HLT) &Mamba  & 90.87 & & 90.53 & & 86.22\\
\rowcolor{gray!15}
\ Ours (SAST) &Mamba  & \textbf{92.25} & & \textbf{91.39} & & \textbf{87.30}\\
 \midrule\midrule
 \multicolumn{7}{c}{\textit{Training from pretrained}}\\
 \midrule

\ Ours (HLT) &Mamba  & 92.94 & & 91.42 & & 87.52\\
 \rowcolor{gray!15}
 \ Ours (SAST) &Mamba  & \textbf{94.32} & & \textbf{91.91} & & \textbf{89.10}\\
  \bottomrule
\end{tabular}
\end{table}

\mypar{Number of Eigenvectors in HLT:}
\cref{tab:scanobjnn_hlt} evaluates the impact of varying the number of eigenvectors in our proposed \gls{hlt} strategy on part segmentation performance using the ShapeNetPart dataset, considering both \textit{training from scratch} and \textit{training from pretrained weights}.

In both scenarios, the segmentation accuracy, measured by the mean Intersection over Union (mIoU), demonstrates that the number of eigenvectors directly influences performance. The accuracy peaks at four eigenvectors, achieving 85.9\% (scratch) and 86.1\% (pretrained), as this setting provides an optimal balance for spatial encoding. 

When the number of eigenvectors is low, the model fails to partition points accurately, resulting in unrelated points being grouped into the same segment. Conversely, when the number of eigenvectors is high, the performance decreases slightly due to redundancy and noise. Higher-order eigenvectors encode finer details or localized variations, which may not align with meaningful segmentation. This can lead to overfitting or confusion between closely related parts.


\begin{table}[h]
  \centering
  \footnotesize	
  \caption{Part segmentation on ShapeNetPart.}
  \setlength{\tabcolsep}{3.5 mm} 
  \label{tab:scanobjnn_hlt}
  \begin{tabular}{lccc}
    \toprule
    Methods & Param. (M) & Eigenvectors & mIoU \\
    \midrule\midrule
    \multicolumn{4}{c}{\textit{Training from scratch}} \\
    \midrule
    Ours (HLT) & 12.3 & 3 & 85.6 \\
    \rowcolor{gray!15} Ours (HLT) & 12.3 & \textbf{4} & \textbf{85.9} \\
    Ours (HLT) & 12.3  & 5 & 85.9 \\
    Ours (HLT) & 12.3 & 6 & 85.8 \\
    \midrule
    \multicolumn{4}{c}{\textit{Training from pretrained}} \\
    \midrule
    Ours (HLT) & 12.3  & 3 & 85.8 \\
    \rowcolor{gray!15} Ours (HLT) & 12.3  & \textbf{4} & \textbf{86.1} \\
    Ours (HLT) & 12.3  & 5 & 86.0 \\
    Ours (HLT) & 12.3  & 6 & 85.8 \\
    \bottomrule
  \end{tabular}
\end{table}

\section{Additional Visualization}

\begin{figure*}[!t]
    \centering
    \includegraphics[width=1.0\textwidth]{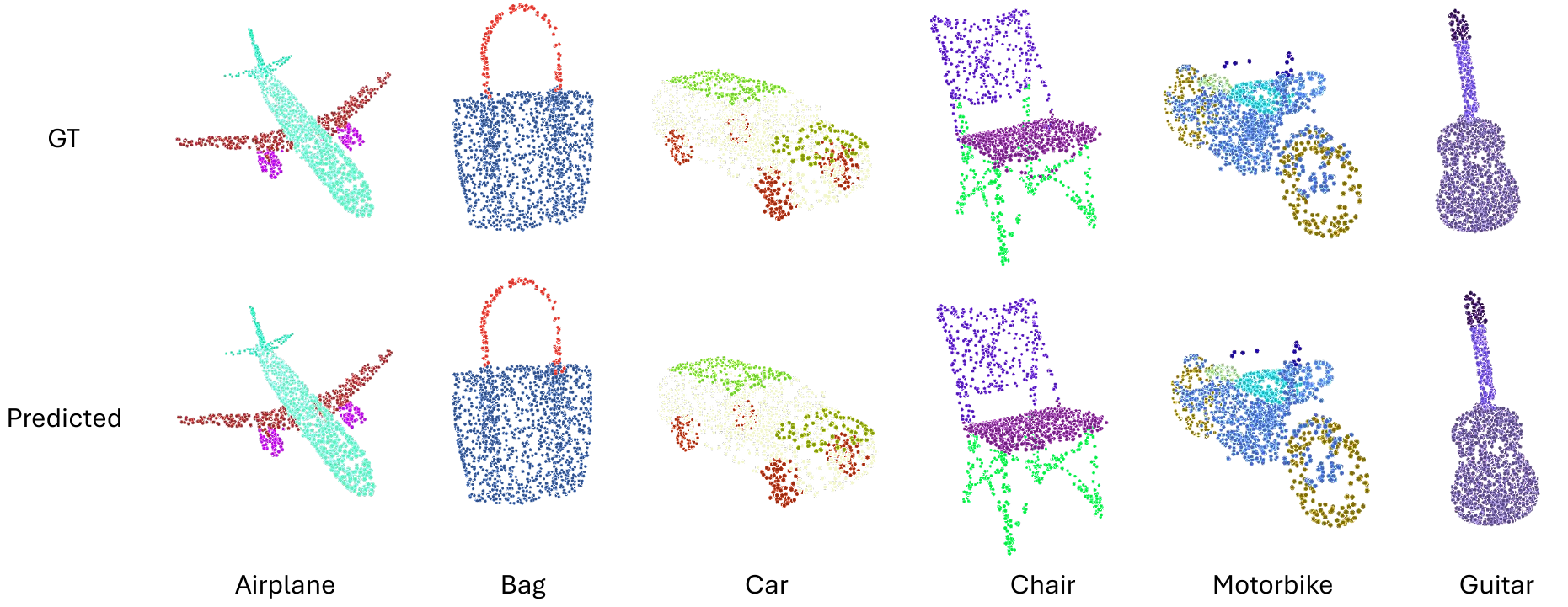}
    \caption{The qualitative results of part segmentation of our HLT method on ShapeNetPart dataset}
    \label{fig:seg}
\end{figure*}

\mypar{Segmentation Results.}
\cref{fig:seg} provides results for six object categories (``Airplane,'' ``Bag,'' ``Car,'' ``Chair,'' ``Motorbike,'' and ``Guitar) obtained by our \gls{hlt} method. In this figure, each point is color-coded based on its class label. The comparison between the ground truth (GT) and the predicted segmentation demonstrates the outstanding performance of our method, as well as its ability to capture fine-grained details.

\begin{figure*}[!t]
    \centering
    \includegraphics[width=0.825\textwidth]{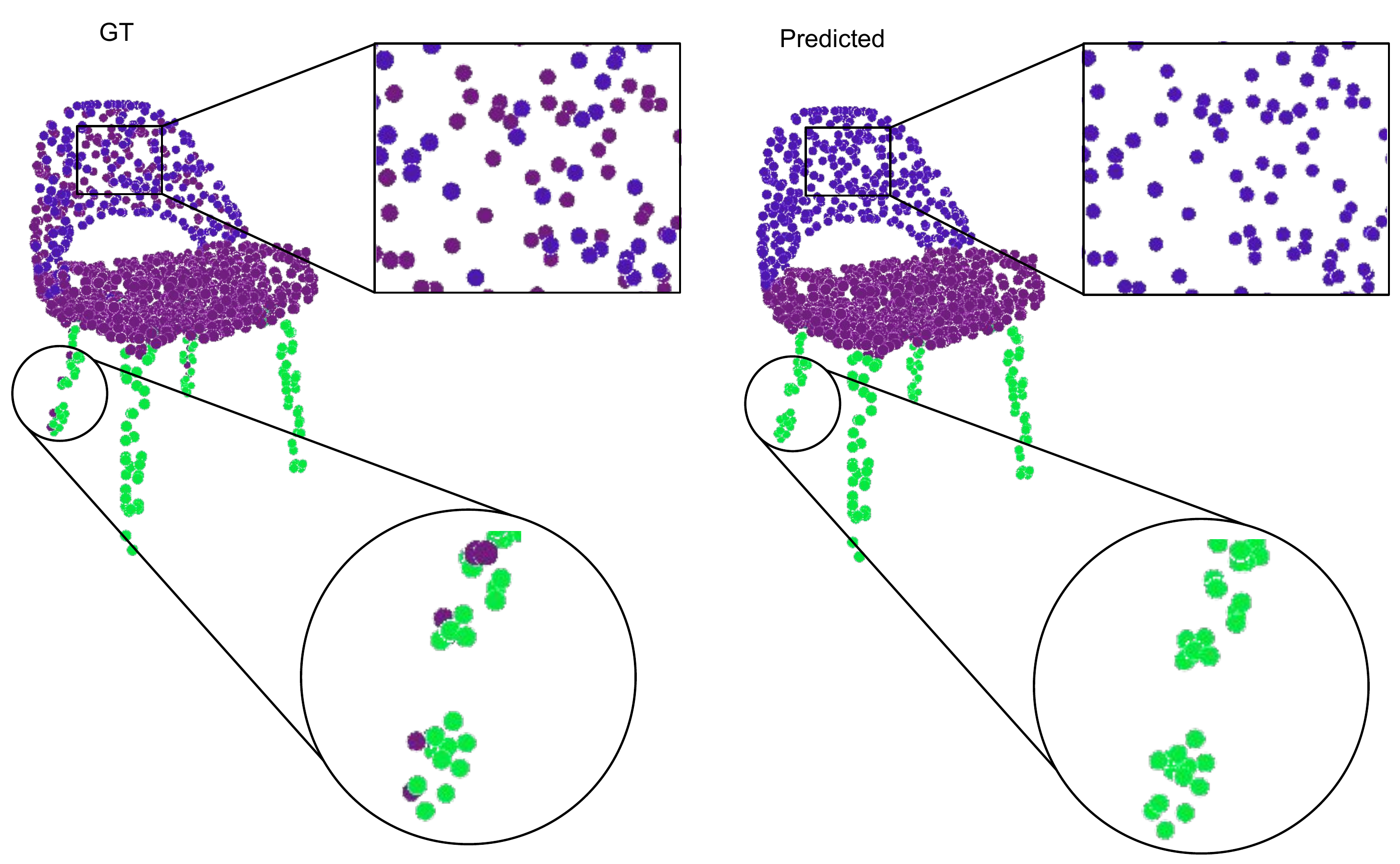}
    \caption{An example illustrating inconsistencies in the ground truth (GT) annotations. The predicted segmentation (right) is geometrically more accurate and consistent compared to the GT (left).}
    \label{fig:seg_bug}
\end{figure*}

\mypar{Dataset Challenges and Ground Truth Anomalies in ShapeNetPart.}
The ShapeNetPart dataset is widely recognized as a challenging benchmark for 3D point cloud segmentation. Upon further investigation of the dataset, we observed certain inconsistencies and inaccuracies in the provided GT annotations. As illustrated in ~\cref{fig:seg_bug}, our method  exhibits a visually better segmentation than the ground truth. 

Such discrepancies in the ground truth highlight potential limitations in the dataset itself, which poses a challenge for both training and evaluation. This phenomenon also provides an explanation for the observation that recent state-of-the-art methods (as listed in Table 2 of the main paper) achieve similar mIOU performance on this dataset, with only marginal improvements. The inherent noise and errors in the ground truth annotations make it difficult for methods to demonstrate significant gains in segmentation quality.

Despite these challenges, our method still achieves competitive performance while maintaining robust predictions that align closely with the underlying geometric features of the objects. 

\begin{figure*}[!t]
    \centering
    \includegraphics[width=1.0\textwidth]{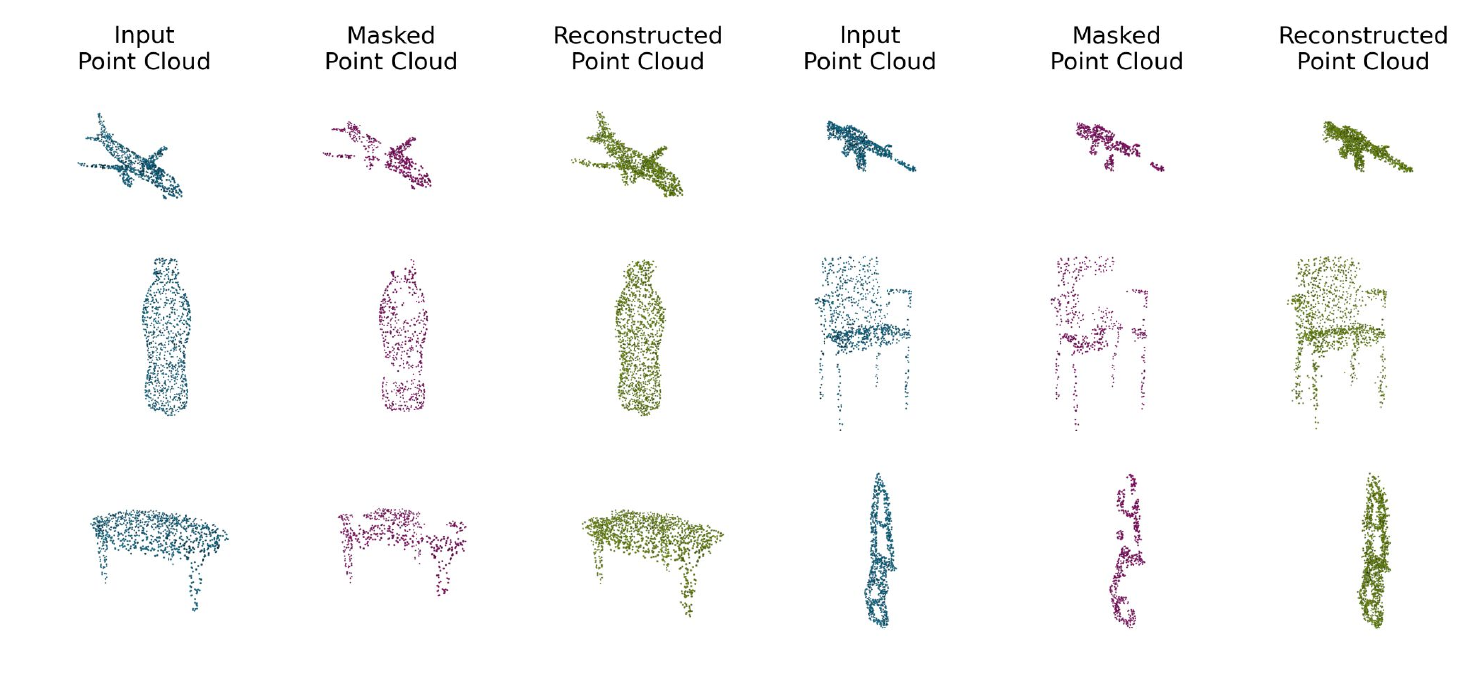}
    \caption{Reconstruction results on the ShapeNet dataset}
    \label{fig:rec}
\end{figure*}

\mypar{Reconstruction Results.}
\cref{fig:rec} showcases the reconstruction capability of Masked Autoencoders (MAEs) on point cloud data using the ShapeNet dataset. The figure consists of three columns for each sample, illustrating the progression from the input point cloud to the final reconstructed result.

The first column, ``Input Point Cloud'', represents the original point cloud data, providing a complete view of the object before any masking or processing. This serves as a reference for evaluating the quality of the reconstruction. The second column, displays the same point cloud after a portion of the data has been masked out. The visible points indicate the sparse information available to the model during the reconstruction phase.
The third column, ``Reconstructed Point Cloud'', demonstrates the MAE's ability to predict and restore the masked regions, resulting in a nearly complete reconstruction that aligns closely with the original structure. 

These results underline the effectiveness of MAEs in capturing and reconstructing geometric details from incomplete data, making them well-suited for extracting meaningful features.

\end{document}